\documentclass[conference]{IEEEtran}
\IEEEoverridecommandlockouts

\usepackage{cite}
\usepackage{amsmath,amssymb,amsfonts}
\usepackage{graphicx}
\usepackage{booktabs}
\usepackage{array}
\usepackage{multirow}
\usepackage{xcolor}
\usepackage{url}
\usepackage[hidelinks,hyperfootnotes=false]{hyperref}
\hypersetup{pdftitle={ScanDreamer: A Map-Grounded World Model for Quadrotor Navigation}}

\newcommand{\ours}{ScanDreamer}
\newcommand{\msd}[2]{#1\,$\pm$\,#2}
\makeatletter
\newcommand{\res}[2]{\@ifundefined{res@#1@#2}{\errmessage{Missing result macro #1/#2}}{\@nameuse{res@#1@#2}}}
\newcommand{\setres}[3]{\@namedef{res@#1@#2}{#3}}
\makeatother
\setres{NAVRLZS}{ID}{28.5}
\setres{NAVRLRT08}{ID}{5.3}
\setres{NAVRLRT20}{ID}{15.1}
\setres{TDMPC208}{ID}{7.3}
\setres{TDMPC210}{ID}{5.6}
\setres{TDMPC220}{ID}{7.4}
\setres{AD08}{ID}{\msd{61.8}{3.4}}
\setres{AD10}{ID}{\msd{62.1}{5.0}}
\setres{AD20}{ID}{\msd{66.0}{10.1}}
\setres{OURS}{ID}{\textbf{\boldmath \msd{97.3}{1.0}}}
\setres{NAVRLZS}{STR}{47.3}
\setres{NAVRLRT08}{STR}{12.3}
\setres{NAVRLRT20}{STR}{5.7}
\setres{TDMPC208}{STR}{15.3}
\setres{TDMPC210}{STR}{11.7}
\setres{TDMPC220}{STR}{9.3}
\setres{AD08}{STR}{\msd{73.7}{2.5}}
\setres{AD10}{STR}{\msd{74.0}{5.8}}
\setres{AD20}{STR}{\msd{74.9}{5.1}}
\setres{OURS}{STR}{\textbf{\boldmath \msd{97.0}{0.3}}}
\setres{NAVRLZS}{OPEN}{\textbf{\boldmath 100.0}}
\setres{NAVRLRT08}{OPEN}{45.0}
\setres{NAVRLRT20}{OPEN}{88.0}
\setres{TDMPC208}{OPEN}{55.0}
\setres{TDMPC210}{OPEN}{48.0}
\setres{TDMPC220}{OPEN}{61.0}
\setres{AD08}{OPEN}{\msd{96.0}{2.6}}
\setres{AD10}{OPEN}{\msd{94.7}{3.8}}
\setres{AD20}{OPEN}{\msd{96.0}{6.1}}
\setres{OURS}{OPEN}{\textbf{\boldmath \msd{100.0}{0.0}}}
\setres{NAVRLZS}{PARAM}{27.9}
\setres{NAVRLRT08}{PARAM}{4.3}
\setres{NAVRLRT20}{PARAM}{12.1}
\setres{TDMPC208}{PARAM}{3.6}
\setres{TDMPC210}{PARAM}{3.6}
\setres{TDMPC220}{PARAM}{5.0}
\setres{AD08}{PARAM}{\msd{44.5}{3.3}}
\setres{AD10}{PARAM}{\msd{43.6}{3.3}}
\setres{AD20}{PARAM}{\msd{47.1}{4.3}}
\setres{OURS}{PARAM}{\textbf{\boldmath \msd{89.5}{1.5}}}
\setres{NAVRLZS}{TOPO}{8.6}
\setres{NAVRLRT08}{TOPO}{0.5}
\setres{NAVRLRT20}{TOPO}{0.5}
\setres{TDMPC208}{TOPO}{2.3}
\setres{TDMPC210}{TOPO}{2.7}
\setres{TDMPC220}{TOPO}{4.1}
\setres{AD08}{TOPO}{\msd{25.3}{3.1}}
\setres{AD10}{TOPO}{\msd{23.3}{5.6}}
\setres{AD20}{TOPO}{\msd{30.9}{12.1}}
\setres{OURS}{TOPO}{\textbf{\boldmath \msd{94.4}{1.6}}}
\setres{NAVRLZS}{SR}{33.8}
\setres{NAVRLRT08}{SR}{8.4}
\setres{NAVRLRT20}{SR}{15.6}
\setres{TDMPC208}{SR}{10.8}
\setres{TDMPC210}{SR}{8.9}
\setres{TDMPC220}{SR}{10.5}
\setres{AD08}{SR}{\msd{59.6}{3.0}}
\setres{AD10}{SR}{\msd{59.4}{4.2}}
\setres{AD20}{SR}{\msd{63.0}{8.5}}
\setres{OURS}{SR}{\textbf{\boldmath \msd{96.3}{0.3}}}
\setres{NAVRLZS}{CR}{11.3}
\setres{NAVRLRT08}{CR}{37.1}
\setres{NAVRLRT20}{CR}{16.7}
\setres{TDMPC208}{CR}{52.1}
\setres{TDMPC210}{CR}{51.9}
\setres{TDMPC220}{CR}{52.5}
\setres{AD08}{CR}{\msd{28.8}{2.8}}
\setres{AD10}{CR}{\msd{29.7}{10.9}}
\setres{AD20}{CR}{\msd{23.1}{5.4}}
\setres{OURS}{CR}{\textbf{\boldmath \msd{3.6}{0.3}}}
\setres{NAVRLZS}{TO}{54.5}
\setres{NAVRLRT08}{TO}{54.5}
\setres{NAVRLRT20}{TO}{67.6}
\setres{TDMPC208}{TO}{37.0}
\setres{TDMPC210}{TO}{39.0}
\setres{TDMPC220}{TO}{36.9}
\setres{AD08}{TO}{\msd{11.3}{0.8}}
\setres{AD10}{TO}{\msd{9.6}{9.5}}
\setres{AD20}{TO}{\msd{12.9}{3.1}}
\setres{OURS}{TO}{\textbf{\boldmath \msd{0.1}{0.0}}}
\setres{NAVRLZS}{SPL}{0.324}
\setres{NAVRLRT08}{SPL}{0.077}
\setres{NAVRLRT20}{SPL}{0.144}
\setres{TDMPC208}{SPL}{0.096}
\setres{TDMPC210}{SPL}{0.076}
\setres{TDMPC220}{SPL}{0.095}
\setres{AD08}{SPL}{\msd{0.496}{0.039}}
\setres{AD10}{SPL}{\msd{0.507}{0.042}}
\setres{AD20}{SPL}{\msd{0.517}{0.089}}
\setres{OURS}{SPL}{\textbf{\boldmath \msd{0.831}{0.008}}}
\setres{OURS}{ASR}{\textbf{\boldmath \msd{96.3}{0.3}}}
\setres{NOCONE}{ASR}{\msd{89.9}{3.1}}
\setres{NOARR}{ASR}{\msd{96.2}{1.6}}
\setres{NOCA}{ASR}{\msd{90.3}{3.3}}
\setres{GEO}{ASR}{\msd{51.6}{2.5}}
\setres{OURS}{ACR}{\msd{3.6}{0.3}}
\setres{NOCONE}{ACR}{\msd{9.9}{3.1}}
\setres{NOARR}{ACR}{\textbf{\boldmath \msd{3.4}{1.6}}}
\setres{NOCA}{ACR}{\msd{9.2}{3.1}}
\setres{GEO}{ACR}{\msd{32.5}{8.8}}
\setres{OURS}{ATO}{\textbf{\boldmath \msd{0.1}{0.0}}}
\setres{NOCONE}{ATO}{\textbf{\boldmath \msd{0.1}{0.1}}}
\setres{NOARR}{ATO}{\msd{0.3}{0.1}}
\setres{NOCA}{ATO}{\msd{0.5}{0.2}}
\setres{GEO}{ATO}{\msd{14.4}{9.7}}
\setres{OURS}{ASPL}{\textbf{\boldmath \msd{0.831}{0.008}}}
\setres{NOCONE}{ASPL}{\msd{0.770}{0.021}}
\setres{NOARR}{ASPL}{\msd{0.815}{0.027}}
\setres{NOCA}{ASPL}{\msd{0.757}{0.029}}
\setres{GEO}{ASPL}{\msd{0.431}{0.034}}
\setres{OURS}{FWD}{\textbf{\boldmath \msd{92.4}{0.9}}}
\setres{NOCONE}{FWD}{\msd{40.1}{7.0}}
\setres{NOARR}{FWD}{\msd{86.4}{2.3}}
\setres{NOCA}{FWD}{\msd{36.4}{7.4}}
\setres{GEO}{FWD}{\msd{48.7}{3.5}}
\setres{OURS}{APPR}{\textbf{\boldmath \msd{0.38}{0.01}}}
\setres{NOCONE}{APPR}{\msd{0.40}{0.04}}
\setres{NOARR}{APPR}{\msd{1.39}{0.30}}
\setres{NOCA}{APPR}{\msd{1.02}{0.27}}
\setres{GEO}{APPR}{\msd{0.99}{0.38}}
\setres{OURS}{OFF}{\msd{35.8}{10.8}}
\setres{NOCONE}{OFF}{\msd{77.0}{3.2}}
\setres{NOARR}{OFF}{\msd{40.2}{9.2}}
\setres{NOCA}{OFF}{\msd{74.2}{13.6}}
\setres{GEO}{OFF}{\msd{83.6}{4.1}}
\setres{NAVRLRT08}{SEEDS}{1}
\setres{NAVRLRT20}{SEEDS}{1}
\setres{TDMPC208}{SEEDS}{1}
\setres{TDMPC210}{SEEDS}{1}
\setres{TDMPC220}{SEEDS}{1}
\setres{AD08}{SEEDS}{3}
\setres{AD10}{SEEDS}{3}
\setres{AD20}{SEEDS}{3}
\setres{OURS}{SEEDS}{3}
\setres{NOCONE}{SEEDS}{3}
\setres{NOARR}{SEEDS}{3}
\setres{NOCA}{SEEDS}{3}
\setres{GEO}{SEEDS}{3}

\begin{document}
\bstctlcite{IEEEexample:BSTcontrol}

\title{ScanDreamer: A Map-Grounded World Model for Quadrotor Navigation}

\author{%
\IEEEauthorblockN{Yunyi Gao$^{1,*}$,
Yixing Shi$^{1,*}$,
Yize Liu$^{1,*}$,
Hai-Duong  Tran$^{2}$,
Nhat-Hoang P. Nguyen$^{3}$,
Shuai Wang$^{1}$, \\
Yi Deng$^{1}$,
Shengming Liu$^{1}$,
Sim Kuan Goh$^{1,\dagger}$,
Duc-Thinh Pham$^{2,3,\dagger}$, and
Sameer Alam$^{3}$}
\IEEEauthorblockA{$^{1}$School of Artificial Intelligence and Robotics, Xiamen University Malaysia, Malaysia\quad \\
$^{2}$Center for AI Research, VinUniversity, Hanoi, Vietnam\quad \\
$^{3}$Air Traffic Management Research Institute, Nanyang Technological University, Singapore\quad\\
}}

\maketitle
\begingroup
\renewcommand{\thefootnote}{}%
\footnotetext{%
\begin{flushleft}
$^{*}$Equal Contribution.\\
$^{\dagger}$Corresponding: simkuan.goh@xmu.edu.my, dtpham@ntu.edu.sg
\end{flushleft}
}%
\endgroup

\begin{abstract}\boldmath
Quadrotors inspecting buildings and tunnels must navigate unknown clutter with a forward-facing depth camera whose limited field of view constrains scene understanding and collision-avoiding navigation. Recent world-model agents learn navigation through imagined experience (e.g., AirDreamer), but remain data-intensive and often lack explicit geometric memory and awareness of what the camera can observe. We propose \ours{}, a map-grounded world model that constructs an online 2D scan-memory map from depth images and vehicle pose. Our world model, actor, and critic are conditioned on this map, which also defines a geodesic progress reward. Moreover, a forward-cone penalty discourages horizontal motion outside the camera's viewing direction. During imagination, \ours{} propagates the quadrotor pose over a frozen map and derives progress and wall-contact signals from the mapped geometry. We train \ours{} for 0.8 million environment steps (5.5\,h on one GPU) and evaluate it in NVIDIA Isaac Sim over 1560 episodes spanning unseen layouts and parameter ranges. Across three seeds (mean\,$\pm$\,SD), \ours{} achieves \msd{96.3}{0.3}\% success with \msd{3.6}{0.3}\% collision, compared with \msd{59.6}{3.0}\% success and \msd{28.8}{2.8}\% collision for AirDreamer at the same online-interaction budget. Removing the forward-cone penalty increases collisions to \msd{9.9}{3.1}\%. These results show that combining explicit map memory, map-grounded imagination, and perception-aware motion substantially improves navigation performance and data efficiency. The source code is available at \url{https://github.com/WANGSHUAIzzzz/ScanDreamer}.
\end{abstract}


\section{Introduction}\label{sec:intro}

\begin{figure*}[t]
  \centering
  \includegraphics[width=\textwidth]{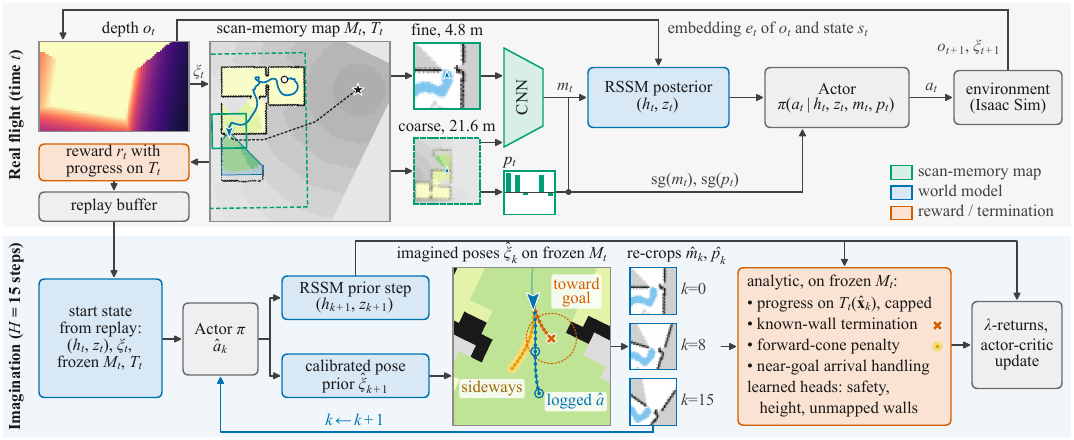}
  \caption{Overview of \ours{}. Top (flight): each depth image
  $o_t$ updates the scan-memory map $M_t$ (black, occupied; gray, unobserved; shading, cost-to-go $T_t$, darker is
  lower), and crops and $T_t$ at the pose $\xi_t$ give the map feature $m_t$ and the plan vector $p_t$. Bottom
  (imagination): the calibrated pose prior converts action sequences from this state into poses $\hat\xi_k$ on the
  frozen map, and the crops and $T_t$ are read again at each $\hat\xi_k$. The blue
  sequence replays the next 15 logged actions. Of the two others, the one toward the goal hits a mapped wall
  ($\times$), and the sideways one incurs the forward-cone penalty (halo).}
  \label{fig:overview}
\end{figure*}

In buildings, tunnels, and caves, where sending people is slow or dangerous, small unmanned aerial vehicles (UAVs)
are used for inspection, search and rescue, and exploration~\cite{tranzatto2022cerberus}. Such missions often lack a
reliable link to a human pilot and rarely come with prior maps, so the vehicle must find its own way through unknown
clutter with its onboard sensors alone. Since a collision usually ends the mission, avoiding one matters as much as
reaching the goal.

Classical systems map, plan, and track in a chain of modules~\cite{zhou2020ego,tordesillas2021faster}.
This modular structure is interpretable, but its objectives and costs
typically require manual design, while errors in perception or mapping can
propagate through the remaining pipeline. Learning-based methods instead
optimize navigation policies through interaction in simulation. Such
policies have enabled high-speed flight through clutter using depth
images~\cite{loquercio2021learning}, autonomous drone racing against human
champions~\cite{kaufmann2023champion}, and avoidance of moving
obstacles~\cite{xu2025navrl}. World-model agents~\cite{hafner2020dream,hafner2025dreamerv3} seek to
reduce environment interaction by learning a predictive model and
training the policy on rollouts imagined within it. On drones,
however, they still train slowly. AirDreamer~\cite{liu2026airdreamer} uses 2.5M environment
steps for depth-based navigation, while Dream to
Fly~\cite{romero2026dream} reports approximately 240\,h of training to
converge on pixel-based drone racing.

Autonomous flight in confined environments is still limited by costly
training, by obstacles forgotten once out of view, and by motion into
poorly observed space. Existing world-model agents do not fully use depth
images and poses to build a geometric memory or account for the direction of
sensing, which leaves them with four limitations. First, a recurrent
memory retains obstacle geometry poorly in large or maze-like
environments~\cite{liu2026airdreamer,romero2026dream,pasukonis2023evaluating}.
Second, Euclidean progress can reward flight into dead ends, while standard
geodesic progress needs ground-truth
geometry~\cite{savva2019habitat,wijmans2020ddppo}, which the vehicle lacks.
Third, if the reward ignores the camera direction, it can encourage sideways
or backward flight into poorly observed space. Fourth, imagined rollouts are purely
latent: the learned heads for progress, wall contacts, and arrival see
neither the imagined pose nor mapped walls, and the actor can exploit
their errors~\cite{ha2018recurrent}.

To address these limitations, we propose \ours{}
(Fig.~\ref{fig:overview}), a map-grounded navigation agent. It builds on
DreamerV3~\cite{hafner2025dreamerv3} and adds one component for each
limitation. We evaluate it in simulation on five test suites of new scenes,
including layouts, parameters, and obstacle cases absent from training,
against AirDreamer, NavRL~\cite{xu2025navrl}, and
TD-MPC2~\cite{hansen2024tdmpc2}. At the same training budget, \ours{} succeeds far more often
than AirDreamer and collides far less often, and it stays ahead when
AirDreamer trains for 2.5 times as many steps. World-model agents for small UAVs
can thus learn safer flight from less data by building on what the vehicle
knows and sees. Because pose estimates and
known sensor orientations are available on many mobile robots, these
principles may also extend beyond quadrotors. Our contributions are as follows:
\begin{itemize}
  \item \textbf{Scan-memory map.} We construct a persistent 2-D occupancy
        map from depth observations and poses, and condition the world
        model, actor, and critic on local map crops
        (Sec.~\ref{sec:map}).

  \item \textbf{Map-grounded imagination.} In imagined rollouts, a
        calibrated pose moves over the frozen map, and progress, contacts
        with mapped walls, and goal arrival are computed analytically
        (Secs.~\ref{sec:imag} and~\ref{sec:ablation}).

  \item \textbf{Map- and perception-aware rewards.} We compute geodesic
        progress from a cost-to-go field over the accumulated map
        (Sec.~\ref{sec:geo}). We also introduce a forward-cone penalty
        that discourages horizontal motion away from the camera axis and
        toward poorly observed space
        (Secs.~\ref{sec:cone} and~\ref{sec:ablation}).
\end{itemize}

\section{Related Work}\label{sec:related}

\subsection{Learning-Based Navigation}

Learning-based methods have been studied across aviation, from
traffic-level decision support to autonomous flight
\cite{pham2019reinforcement,tran2020interactive,
loquercio2021learning,song2023learning}. Depth-based flight policies have
been learned by imitating privileged experts
\cite{loquercio2021learning,song2023learning} and through differentiable
dynamics~\cite{zhang2025learning,lee2025depthnav}. Model-free
NavRL~\cite{xu2025navrl} uses PPO~\cite{schulman2017ppo} to train a policy
that reads distances ray-cast against a map rather than depth images.
Perception-aware methods keep a point of interest in
view~\cite{falanga2018pampc}, restrict commanded velocity to the sensor's
field of view~\cite{kulkarni2024reinforcement}, reward yaw alignment with
a privileged reference path~\cite{song2023learning}, or align the body
$x$-axis with a smoothed velocity~\cite{lee2025depthnav}. These approaches
either constrain motion or rely on privileged reference information.
This motivates perception-aware guidance that favors motion toward
observed space while preserving the full action space and requiring
neither a reference path nor a heading target.

\subsection{World Models for Robot Learning and Flight}

Learned predictive models have been applied across flight and robot
learning~\cite{hafner2020dream,hafner2025dreamerv3,
goh2021tunnel,goh2023unstable}. Dreamer agents have been applied to depth-based drone
navigation~\cite{liu2026airdreamer} and drone racing from
pixels~\cite{romero2026dream}. Whereas these agents act using policies
learned through imagination, TD-MPC2~\cite{hansen2024tdmpc2}, a
planning-based method, selects each action using model predictive path
integral control in the latent space of a decoder-free world model. In
concurrent work, MAD~\cite{zhang2026mad} decodes robocentric occupancy
and visibility maps from its latent state, but its imagined rollouts
remain latent. DreamerNav~\cite{shanks2025dreamernav}, a DreamerV3 agent
for ground robots, observes a local occupancy map with an A* route planned
on a known static map. Existing methods therefore either keep imagination
purely latent or depend on a known map. A map constructed online offers a
natural way to ground imagined rollouts, combining learned prediction
with explicit geometric reasoning.

\subsection{Spatial Memory and Progress Rewards}

Spatial reasoning shapes decision-making in both human-operated and
autonomous systems~\cite{zhong2022spatial,chaplot2020learning,
pasukonis2023evaluating}. Active Neural SLAM
\cite{chaplot2020learning} constructs a top-down metric map online and
guides a learned local policy using a planner over this map.
MAVRL~\cite{yu2025mavrl} retains past quadrotor depth observations in a
learned latent representation rather than an explicit map.
R2I~\cite{samsami2024mastering} strengthens latent memory using
state-space models and, after 400M environment steps, exceeds human scores
on all but the largest mazes in Memory Maze
\cite{pasukonis2023evaluating}. Point-goal progress is commonly measured using geodesic distance over
ground-truth meshes~\cite{savva2019habitat,wijmans2020ddppo} or
time-of-arrival maps of the training environments
\cite{lee2025depthnav}. These approaches provide accurate training
signals but depend on privileged knowledge of the environment. The
agent's partial online map can instead serve two roles: preserving
previously observed geometry and providing a measure of
geodesic progress.
\section{Method}
\label{sec:method}

\subsection{Problem Formulation and Base Agent}
\label{sec:problem}

We formulate point-goal navigation without a prior map as a partially observable Markov decision process. An
episode is truncated at the step limit of its scene and terminates earlier on arrival, collision, or exit from the
altitude band of 0.3 to 3.2\,m. The quadrotor arrives when its 3D distance $d^{\mathrm{3D}}_t$ to the
goal falls below $r_g=1$\,m and collides when its horizontal distance $d_{\mathrm{obs}}$ to the nearest obstacle falls
below 0.3\,m.

At each step, the agent observes a $48\times80$ depth image $o_t$ from a forward-facing camera with an $87^\circ$
horizontal FOV and a range of 0.6 to 6\,m. A 15-D state vector $s_t$ holds the goal direction and
distances, the velocity, the body rates, and the attitude. Each action $a_t\in[-1,1]^5$ is applied at 20\,Hz. Its
first three components are a velocity setpoint in the yaw-aligned heading frame, scaled by
$\Lambda=\mathrm{diag}(1.5,1.5,0.5)$\,m/s, followed by the cosine and sine of an absolute yaw command. To build its
map, \ours{} also reads its height above the ground and the planar pose $\xi_t=(\mathbf{x}_t,\psi_t)$ in the take-off
frame.

\ours{} builds on DreamerV3~\cite{hafner2025dreamerv3}, using its
recurrent state-space model (RSSM)~\cite{hafner2019learning} and
actor--critic training. The actor $\pi_\theta$ and critic $V_\phi$ learn
from $\lambda$-returns over $H=15$ imagined steps. As in
AirDreamer~\cite{liu2026airdreamer}, the actor loss includes an
action-smoothness term. At each step, the agent receives
\begin{equation}
  r_t = r^{\mathrm{geo}}_t + r^{\mathrm{fwd}}_t
        + 5\,r^{\mathrm{safe}}_t + 2\,r^{\mathrm{hgt}}_t
        + r^{\mathrm{goal}}_t + r^{\mathrm{col}}_t ,
  \label{eq:reward}
\end{equation}
where $r^{\mathrm{geo}}_t$ replaces AirDreamer's Euclidean progress
reward and $r^{\mathrm{fwd}}_t$ penalizes motion outside the forward
cone. The remaining terms follow AirDreamer:
$r^{\mathrm{safe}}_t=-\max(0,1-d_{\mathrm{obs}})^2$ penalizes obstacle
proximity, and $r^{\mathrm{hgt}}_t=-\Delta z_t^2$ penalizes the vertical
goal offset. Arrival gives $r^{\mathrm{goal}}_t=100$, while collision
gives $r^{\mathrm{col}}_t=-5$.

\subsection{Scan-Memory Map}
\label{sec:map}

The scan-memory map $M_t$ (Fig.~\ref{fig:overview}) is a persistent occupancy map in the take-off frame, built
online from depth and pose, and retains obstacles outside the camera's view. Depth pixels are back-projected with
the attitude, and those 0.25 to 3.4\,m above ground are taken as obstacle points. The nearest obstacle point per
image column terminates a horizontal ray. Rays update a 0.15\,m grid with clamped
log-odds~\cite{hornung2013octomap}, and occupied cells are never cleared during an episode. Out-of-range pixels are
set to 6\,m, and 6\,m readings (either free space or a wall within 0.6\,m) are discarded.

The geodesic cost-to-go $T_t$ is computed every 10 steps and at any step that adds occupied cells, using Dijkstra's
algorithm on 0.3\,m nodes of $M_t$, where unknown cells are traversable. Nodes within 0.3\,m of an occupied cell are
blocked, and the node cost drops from 3 at 0.3\,m clearance to 1 at 0.8\,m, so $T_t$ equals the path length where
the clearance is at least 0.8\,m and exceeds it near walls.

Map patches are stored with every step, and the networks read heading-aligned crops of them at the pose $\xi_t$.
The fine crop covers $32\times32$ cells of 0.15\,m and marks occupied, known, and visited cells. The coarse crop
($36\times36$ cells of 0.6\,m) holds the occupied and known fractions, $T_t$ relative to its value at the pose, and a
goal mask. A CNN encodes the two crops into a 256-D map feature $m_t$. A 6-D plan vector
$p_t$ holds $T_t$, its descent direction in the body frame, a detour ratio, a visited flag, and $\Delta z_t$.

In the world model, $m_t$ enters the posterior with $e_t$ but not the prior. The reward and continue heads condition
on $[h_t,z_t,m_t]$, and the actor and critic on $[h_t,z_t,\allowbreak\mathrm{sg}(m_t),\allowbreak\mathrm{sg}(p_t)]$,
with $\mathrm{sg}$ the stop-gradient. A second decoder makes the latent state carry the local map by reconstructing
the occupied and known channels of the fine crop from $(h_t,z_t)$.

\subsection{Geodesic Progress Reward}
\label{sec:geo}

To avoid rewarding progress into dead ends, we compute progress on the geodesic cost-to-go $T_t$ of the quadrotor's
own map (Sec.~\ref{sec:map}) instead of on the straight-line distance. The potential
\begin{equation}
  \Phi_t=\max\Big(\sqrt{T_t(\mathbf{x}_t)^2+\Delta z_t^2}-r_g,\;0\Big)
  \label{eq:potential}
\end{equation}
adds the vertical offset $\Delta z_t$ to it, is clamped at the arrival radius, and is set to zero on the arrival
step. We reward the decrease of this potential, capped by the distance flown,
% \begin{equation}
  $r^{\mathrm{geo}}_t=w_{\mathrm{geo}}\min\big(\Phi_{t-1}-\Phi_t,\;\rho\,\delta_t+\epsilon\big)$,
%   \label{eq:geo}
% \end{equation}
where $\delta_t$ is the 3D step length and $w_{\mathrm{geo}}=10$. In the cap, which leaves negative progress untouched, $\rho=3$ is the largest node cost of $T_t$ and $\epsilon=0.05$\,m is a
small slack. When a newly observed wall raises $T_t$, the step that reveals it is penalized, because $\Phi_{t-1}$
is evaluated on the previous map; flying into a dead end and out again earns no net positive reward.

\subsection{Forward-Cone Penalty}
\label{sec:cone}

To keep the quadrotor from flying into obstacles outside its camera's field of view, we penalize horizontal motion that
leaves a cone around the camera axis. Let $(v_f,v_l)$ be the horizontal velocity in the heading frame, $v$ its magnitude,
and $\beta=\operatorname{atan2}(|v_l|,v_f)$ its bearing from the camera axis. The penalty is
\begin{align}
  r^{\mathrm{fwd}}_t &= -w_{\mathrm{fwd}}\,\Delta t\;g(v_t)\,e_{\mathrm{cone}}(v_t,\beta_t)\,f(d^{\mathrm{3D}}_t),
  \label{eq:cone}\\
  e_{\mathrm{cone}}(v,\beta) &=
  \begin{cases}
    0, & \beta\le\theta_0,\\
    v\sin(\beta-\theta_0), & \theta_0<\beta\le\theta_0+90^\circ,\\
    v, & \text{otherwise},
  \end{cases}
  \label{eq:econe}
\end{align}
where $e_{\mathrm{cone}}$ is the velocity component normal to the edge of a cone with half-angle $\theta_0=20^\circ$,
or the full speed beyond $90^\circ$ from that edge. We scale the weight
$w_{\mathrm{fwd}}=10$ by the control period $\Delta t=0.05$\,s. A gate
$g(v)=\operatorname{clip}((v-v_{\mathrm{rot}})/v_{\mathrm{rot}},0,1)$ with $v_{\mathrm{rot}}=0.15$\,m/s exempts
motion below $v_{\mathrm{rot}}$, such as rotation in place, and near the goal a fade $f(d^{\mathrm{3D}}_t)$ lowers
the weight from full at 1.75\,m to zero at 1.25\,m.

We set $\theta_0$ well below the $43.5^\circ$ half-angle of the FOV. Each action component is clipped separately,
so oblique commands are faster: under the velocity model of Sec.~\ref{sec:imag}, a full command settles at
1.317\,m/s along the heading but at 1.521\,m/s at $30^\circ$ off it. Within the cone, a policy therefore tends to fly
along its edge, which must stay well inside the FOV.

\subsection{Map-Grounded Imagination}
\label{sec:imag}

The scene is static, so an imagined action changes the pose but not the map. \ours{} computes progress, contacts
with mapped walls, and goal arrival from this map at the imagined pose, and the forward-cone penalty from the
imagined velocity, instead of learning them (Fig.~\ref{fig:overview}). The world model still learns the latent dynamics, the safety and height terms, and
contacts with unmapped walls.

\subsubsection{Calibrated Pose Prior} From the replayed state, the pose prior passes the imagined action $\hat a_k$
through first-order lags to the horizontal velocity $\hat{\mathbf{v}}_k$ and the yaw rate $\hat\omega_k$,
\begin{align}
  \hat{\mathbf{v}}_{k+1} &= \hat{\mathbf{v}}_k
    + \alpha_v\big(\mu_v\,\mathbf{R}(\hat\psi_k)\,\Lambda_{xy}\,\hat a^{xy}_k - \hat{\mathbf{v}}_k\big),
  \label{eq:prior}\\
  \hat\omega_{k+1} &= \hat\omega_k + \alpha_\omega\big(\mu_\omega \sin\Delta\psi_k - \hat\omega_k\big),
  \label{eq:yawrate}
\end{align}
where $\Lambda_{xy}$ is the horizontal block of $\Lambda$, $\mathbf{R}(\hat\psi_k)$ rotates the command into the
take-off frame, and $\Delta\psi_k$ is the wrapped error to the commanded yaw. Euler steps
advance the pose, $\hat{\mathbf{x}}_{k+1}=\hat{\mathbf{x}}_k+\hat{\mathbf{v}}_{k+1}\Delta t$ and
$\hat\psi_{k+1}=\hat\psi_k+\hat\omega_{k+1}\Delta t$. A model of the same form gives $\Delta\hat z_k$. We fit both
models on half of 291 episodes from AirDreamer's training replay. On the other half, the median 15-step error of the pose prior is 0.143\,m in position and $3.2^\circ$ in yaw, against 0.363\,m and
$21.3^\circ$ for an uncalibrated kinematic model.

\subsubsection{Re-Cropped Map and Analytic Progress} A rollout keeps the map $M_t$ and the cost-to-go $T_t$ of its
replayed start state fixed. At every imagined pose $\hat\xi_k=(\hat{\mathbf{x}}_k,\hat\psi_k)$, we again crop the map
patches stored with that state, as in Sec.~\ref{sec:map}, to obtain $\hat m_k$ and $\hat p_k$, with no gradient
through the pose. Progress uses $\hat\Phi_k$, the potential~\eqref{eq:potential} at $\hat{\mathbf{x}}_k$ and
$\Delta\hat z_k$, and is
\begin{equation}
  \hat r^{\mathrm{geo}}_k = \mathrm{clip}\big(w_{\mathrm{geo}}(\hat\Phi_{k-1}-\hat\Phi_k),\,-b_k,\,b_k\big),
  \label{eq:imgprog}
\end{equation}
where $b_k=w_{\mathrm{geo}}\,\rho\,(\hat\delta_k+\epsilon')$ caps the progress in both directions by the 3D length
$\hat\delta_k$ of the imagined step, with $\epsilon'=0.02$\,m.

\subsubsection{Known-Wall Termination} Contact with a mapped wall ends an imagined rollout, as a crash ends an episode.
Let $d_k$ be the distance from $\hat{\mathbf{x}}_k$ to the nearest occupied cell of the frozen map. The
hit probability is $\eta_k=\sigma\big((r_c-d_k)/\tau\big)$, with $\tau=0.05$\,m and a contact radius $r_c=0.3$\,m
equal to the collision distance. The cumulative hit $\bar\eta_k=\max_{j\le k}\eta_j$, with $\eta_0=0$, gives the
conditional termination
% \begin{equation}
  $\kappa_k=\frac{\bar\eta_k-\bar\eta_{k-1}}{1-\bar\eta_{k-1}}$,
%   \label{eq:wall}
% \end{equation}
which scales the continue flag $\hat c_k$ by $1-\kappa_k$ and turns the learned reward $\hat r^{\mathrm{wm}}_k$ into
$(1-\kappa_k)\,\hat r^{\mathrm{wm}}_k+\kappa_k\min(\hat r^{\mathrm{wm}}_k,r_{\mathrm{crash}})$, where
$r_{\mathrm{crash}}=-7.45$ is $r^{\mathrm{col}}$ plus the safety term at contact.

\subsubsection{Near-Goal Arrival Handling} Arrival is a sparse event at the edge of the arrival sphere. If the reward
and continue heads must learn it, the critic can value the region just outside the sphere above the arrival reward
itself, and a policy that follows this critic may stall near the goal (Sec.~\ref{sec:ablation}). Arrival handling has
three parts. First, the earliest imagined step $k^\star$ with $\hat d^{\mathrm{3D}}_k<r_g$ ends the rollout and,
barring a known-wall hit on the same step, receives $r^{\mathrm{goal}}=100$ plus the capped progress to a zero
potential. Second, we train the reward head without the
arrival reward and the continue head with arrival counted as continuation, and in imagination we clip the learned
reward at zero, since every term left to it is non-positive. Third, the bootstrap value and the advantage baseline in
imagination are capped by
% \begin{equation}
  $\bar V_k=\gamma^{\max(n_k-1,\,0)}\,r^{\mathrm{goal}}+w_{\mathrm{geo}}\hat\Phi_k$, 
%   \label{eq:valuecap}
% \end{equation}
where $\gamma=1-1/333$ is the discount and $n_k$ is the smallest number of steps that covers the gap
$\max(\hat d^{\mathrm{3D}}_k-r_g,0)$ at the speed bounds of the pose prior, 1.862\,m/s horizontally and 0.46\,m/s
vertically. Because the progress terms telescope, the other terms are non-positive, and the continue flag is
clamped at $\gamma$, $\bar V_k$ bounds the imagined return up to a small slack.

\section{Experiments}
\label{sec:experiments}

\begin{table}[t]
\centering
\caption{Training Set and Test Suites}
\label{tab:suites}
\footnotesize
\setlength{\tabcolsep}{2.5pt}
\begin{tabular}{lrl}
\toprule
Training family & Scenes & Key parameters (training intervals) \\
\midrule
Pillar forest    & 1{,}440 & density 0.02--0.10\,m$^{-2}$, gap 1.0--2.0\,m \\
Wall field       & 1{,}440 & 1--8 walls of 3--10\,m, slits 1.2--1.6\,m \\
Traps            & 2{,}880 & 1--3 U, C, V, or L traps, depth 2.0--4.0\,m \\
Combination      & 4{,}800 & pillar forest, 1--2 traps, 0--3 walls \\
Corridor tree    & 6{,}240 & 2--3 branches, arms 3--9\,m, width 2.1--2.5\,m \\
Grid maze        & 2{,}880 & 5$\times$5 to 7$\times$7 cells of 2.6--3.5\,m \\
Multi-room       & 2{,}400 & rooms 4--7\,m, doors 1.2--1.6\,m \\
Dead-end cluster & 1{,}920 & 3--5 corridors of 3.5--7\,m, one open \\
\bottomrule
\end{tabular}\\[3pt]
\begin{tabular}{lrcccc}
\toprule
Suite & Scenes & Families & Median $R_g$ & Mean $L^*$ (m) & Limit (s) \\
\midrule
In-distribution & 800      & 8  & 1.29 & 21.7 & 60--150 \\
Parameter shift & 140      & 7  & 1.47 & 24.7 & 60--150 \\
Topology shift  & 220      & 10 & 2.74 & 28.0 & 60--150 \\
Structured      & 300      & 6  & 1.26 & 12.4 & 60--69 \\
Open space      & 100      & 1  & 1.00 & 16.5 & 60--64 \\
\bottomrule
\end{tabular}
\end{table}

\subsection{Experimental Setup}
\label{sec:setup}

\subsubsection{Simulator}
We train and test the agents in NVIDIA Isaac Sim with OmniDrones~\cite{xu2024omnidrones} and its low-level position
controller. \ours{}, its variants, and AirDreamer run in Isaac Sim 4.1, and NavRL and TD-MPC2 in Isaac Sim 6.2 with
the same scenes and protocol.
Sixteen quadrotors fly in separate arenas with a 200\,Hz physics step.

\subsubsection{Benchmark Scenes}
Scenes are 24\,m $\times$ 24\,m arenas closed by an outer wall, and their obstacles are vertical boxes 3.5\,m high
that span the flight band. Clearance to the nearest obstacle is at least 1.0\,m at the start and 1.5\,m at the goal,
and every scene admits a collision-free path. A grid planner gives the shortest-path length $L^*$, and the detour
ratio is $R_g=L^*/d_e$ with the start-goal distance $d_e$. The 24,000 training scenes, 28.1\% of them with $R_g>2$,
come from eight of our procedural families (pillar forest to dead-end cluster), and each family generator samples
its layout parameters from fixed intervals (Table~\ref{tab:suites}).

\subsubsection{Test Suites}
\begin{figure}[t]
\centering
\includegraphics[width=\columnwidth]{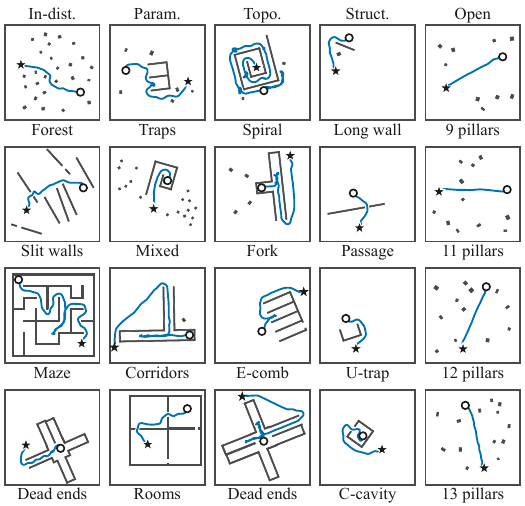}
\caption{Representative top-down test scenes and successful \ours{} trajectories
(blue) from the start (circle) to the goal (star). Columns denote the five
test suites and rows denote scene families. In-dist.\ and Param.\ cover
all eight training families, whereas Struct.\ and Topo.\ show four of six
and four of ten families, respectively. Mixed combines forests, walls,
and traps; Fork is a T-junction with a misleading goal. For each family,
we show the successfully solved scene whose detour ratio $R_g$ is closest
to the family median. Because $R_g\approx1$ throughout open space, those
rows instead show increasing pillar counts.}
\label{fig:suites}
\end{figure}

Whereas the training set is organized by scene family, the five test suites of Table~\ref{tab:suites} are organized
by how they differ from it, and Fig.~\ref{fig:suites} shows typical scenes of each. All are drawn from the same
generator with seeds of their own, so no test scene appears in the training set. The suites form
two groups: the first three move away from the training distribution step by step, and the last two are diagnostic.
The in-distribution suite draws 100 new scenes of each training family from the training intervals.
The parameter-shift suite draws 20 scenes of each of seven training families with every key parameter outside its
training interval, such as longer walls, narrower slits and doors, deeper traps, and denser forests; the dead-end
cluster appears in shifted form in the topology-shift suite, since its shift also moves the goal behind a dead end. This suite changes the layout itself: nine of its ten
families, such as spirals, nested pockets, zigzag dead ends, and a ring corridor whose short route is blocked out
of view, come from generators unused in training, and the tenth is this shifted cluster, with more and longer
corridors and a misleading goal.
These layouts call for memory: on 100 scenes of each family, a planner that keeps only the last 3.2\,s of observations solves at most 49\%,
whereas the same planner with the full scanned map solves all of them. Along these three suites, the median
$R_g$ rises from 1.29 to 1.47 and 2.74 (Table~\ref{tab:suites}).

The first diagnostic suite, the structured suite, isolates six canonical situations absent from training, each a
family of 50 scenes whose parameters vary within narrow ranges and whose start-goal distances are about 6 to 10\,m: a long
wall whose ends lie outside the initial view, a narrow passage with a single gap of 1.0 to 1.15\,m, a U-shaped trap,
a C-shaped cavity in which the quadrotor starts facing the back wall, and two pillar forests with the take-off
heading toward or away from the goal. The second, the open-space suite, holds 100 scenes of 6 to 16 sparse pillars
with almost no detours and serves as a control.

\subsubsection{Training}
\ours{} and its variants use the 100M model size of \mbox{DreamerV3}~\cite{hafner2025dreamerv3}, a train ratio of 64,
and a replay buffer of $3\times10^5$ steps. They train from scratch for 0.8M environment steps, each a control step of
one quadrotor, under dynamics randomized within the ranges of AirDreamer~\cite{liu2026airdreamer}. We
use three seeds (0-2) for \ours{} and its variants as well as for AirDreamer. Tables~\ref{tab:main}
and~\ref{tab:ablation} report the mean and standard deviation (SD), and episode counts in the text are
summed over them. Reaching 0.8M environment steps on one NVIDIA RTX 4090 takes \ours{} 5.5\,h and AirDreamer 5.0\,h.

\subsubsection{Baselines}
Our main baseline is AirDreamer~\cite{liu2026airdreamer}. It reads the same depth image and state as \ours{} but no
pose or map, and we train it with the same action space, training scenes, domain randomization, model size, and
training settings. We test it at our budget of 0.8M steps and after 1.0M and 2.0M steps.

We test the released NavRL policy~\cite{xu2025navrl} with unchanged weights and the rules of its deployment code. It
reads range rays cast on our scan-memory map, where unobserved cells count as free, its yaw command stays on the
start-goal line, and its commands of up to 2\,m/s per axis are clipped to our limits. We also retrain NavRL on our
scenes with its network, reward, and PPO settings and with its actions scaled to our limits, and test it under the
same rules.

TD-MPC2~\cite{hansen2024tdmpc2} uses its 5M configuration with our depth encoder and its recommended settings for
tasks with terminations. It receives three stacked depth frames, AirDreamer's state, and the previous action, and is
trained with AirDreamer's reward. We train NavRL and TD-MPC2 for 0.8M steps and continue NavRL to 2.0M and TD-MPC2
to 1.0M and 2.0M steps. Within these budgets, neither reaches 20\% success on 80 validation
episodes: the retrained NavRL solves 5 and 12 of them after 0.8M and 2.0M steps, and TD-MPC2 solves 7, 6, and 6
after 0.8M, 1.0M, and 2.0M steps.

\newcommand{\mainrow}[1]{\res{#1}{ID} & \res{#1}{PARAM} & \res{#1}{TOPO} & \res{#1}{STR} & \res{#1}{OPEN}
  & \res{#1}{SR} & \res{#1}{CR} & \res{#1}{TO} & \res{#1}{SPL}}
\begin{table*}[t]
\centering
\caption{Comparison with Baselines on the Five Test Suites}
\label{tab:main}
\footnotesize
\setlength{\tabcolsep}{3.1pt}
\begin{tabular}{lcc ccccc cccc}
\toprule
 & & & \multicolumn{5}{c}{SR per suite (\%)$\uparrow$} & \multicolumn{4}{c}{All suites pooled} \\
\cmidrule(lr){4-8}\cmidrule(lr){9-12}
Method & Steps & Seeds & In-dist. & Param. & Topo. & Struct. & Open & SR$\uparrow$ & CR$\downarrow$ & TO$\downarrow$
  & SPL$\uparrow$ \\
\midrule
NavRL~\cite{xu2025navrl}, released & --   & --                     & \mainrow{NAVRLZS} \\
NavRL, retrained                    & 0.8M & \res{NAVRLRT08}{SEEDS} & \mainrow{NAVRLRT08} \\
NavRL, retrained                    & 2.0M & \res{NAVRLRT20}{SEEDS} & \mainrow{NAVRLRT20} \\
TD-MPC2~\cite{hansen2024tdmpc2}     & 0.8M & \res{TDMPC208}{SEEDS}  & \mainrow{TDMPC208} \\
TD-MPC2                             & 1.0M & \res{TDMPC210}{SEEDS}  & \mainrow{TDMPC210} \\
TD-MPC2                             & 2.0M & \res{TDMPC220}{SEEDS}  & \mainrow{TDMPC220} \\
AirDreamer~\cite{liu2026airdreamer} & 0.8M & \res{AD08}{SEEDS}      & \mainrow{AD08} \\
AirDreamer                          & 1.0M & \res{AD10}{SEEDS}      & \mainrow{AD10} \\
AirDreamer                          & 2.0M & \res{AD20}{SEEDS}      & \mainrow{AD20} \\
\midrule
Ours                                & 0.8M & \res{OURS}{SEEDS}      & \mainrow{OURS} \\
\bottomrule
\end{tabular}\\[4pt]
\parbox{\textwidth}{\footnotesize SR, CR, and TO in percent of the suite or of all 1560 test episodes. Bold: best
mean per column. Mean\,$\pm$\,SD over seeds, or a single run where no SD is given.}
\end{table*}

\newcommand{\ablrow}[1]{\res{#1}{ASR} & \res{#1}{ACR} & \res{#1}{ATO} & \res{#1}{ASPL}
  & \res{#1}{FWD} & \res{#1}{OFF} & \res{#1}{APPR}}
\begin{table*}[t]
\centering
\caption{Ablation of \ours{} (0.8M Environment Steps)}
\label{tab:ablation}
\footnotesize
\setlength{\tabcolsep}{3.0pt}
\begin{tabular}{l ccccc cccc ccc}
\toprule
Variant & Map & Geo. & Cone & Imag. & Arr. & SR$\uparrow$ & CR$\downarrow$ & TO$\downarrow$ & SPL$\uparrow$
  & Fwd.\ (\%)$\uparrow$ & Off-axis (\%) & Appr.\ (s)$\downarrow$ \\
\midrule
Ours                 & \checkmark & \checkmark & \checkmark & \checkmark & \checkmark & \ablrow{OURS} \\
w/o forward cone     & \checkmark & \checkmark &            & \checkmark & \checkmark & \ablrow{NOCONE} \\
w/o arrival handling & \checkmark & \checkmark & \checkmark & \checkmark &            & \ablrow{NOARR} \\
w/o cone and arrival & \checkmark & \checkmark &            & \checkmark &            & \ablrow{NOCA} \\
Geodesic reward only &            & \checkmark &            &            &            & \ablrow{GEO} \\
\bottomrule
\end{tabular}\\[4pt]
\parbox{\textwidth}{\footnotesize Units and mean\,$\pm$\,SD as in Table~\ref{tab:main}. Off-axis: off-axis share of
all collisions. Bold: best mean per arrowed column.}
\end{table*}

\subsubsection{Metrics}
An episode ending within 1\,m of the goal counts toward the success rate (SR). Episodes ending in a collision or
at the step limit define the collision rate (CR) and the timeout rate (TO), and all others leave the altitude band.
Success weighted by path length (SPL)~\cite{anderson2018evaluation} uses $L^*$ as the reference length. Over all
successful episodes, the forward-motion share (Fwd.)\ is the fraction of the distance flown that lies within
$\pm30^\circ$ of the camera axis. We call a collision off-axis if its last motion points more than
$30^\circ$ off the camera axis while the speed and the closing speed toward the obstacle exceed 0.2\,m/s. The
final-approach time (Appr.)\ is the time from first coming within 1.5\,m of the goal to arrival. A near-goal stall is
a timeout that ends within 2\,m of the goal.

\subsubsection{Protocol}
All policies fly the same 1560 test episodes under nominal dynamics. Each acts with its mean action, except TD-MPC2,
whose MPPI planner samples its actions. We pair each seed of an agent with the same seed of the agent it is compared
with, and a single-run agent with each seed of the other. A difference is reported only if it points the same way in
every seed pair. A paired count is the number of episodes in which only one of the two policies reaches a given
outcome, such as success.

\subsection{Comparison with Baselines}
\label{sec:main}

\ours{} succeeds in \msd{96.3}{0.3}\% of the test episodes and collides in \msd{3.6}{0.3}\%, whereas AirDreamer
succeeds in \msd{59.6}{3.0}\% and collides in \msd{28.8}{2.8}\% at the same budget (Table~\ref{tab:main}). \ours{}
also succeeds more often on every suite, and training AirDreamer for up to 2.0M steps, 2.5 times our budget, does
not close this gap.

\begin{figure}[t]
\centering
\includegraphics[width=\columnwidth]{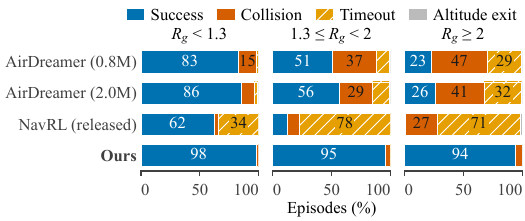}
\caption{Outcomes of the 1560 test episodes, binned by detour ratio $R_g$ (770, 390, and 400 per bin). Segments below
15\% are not labeled.}
\label{fig:detour}
\end{figure}

\begin{figure}[t]
\centering
\includegraphics[width=\columnwidth]{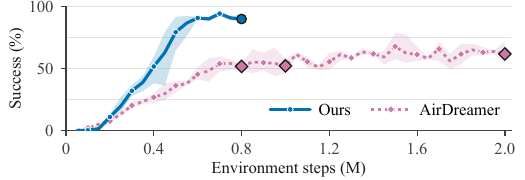}
\caption{Success rate of training episodes in windows of 50k environment steps, shown as the mean (line) and range
(band) over training runs. Black-edged markers are the checkpoints tested in Table~\ref{tab:main}.}
\label{fig:curves}
\end{figure}

\begin{figure*}[t]
\centering
\includegraphics[width=\textwidth]{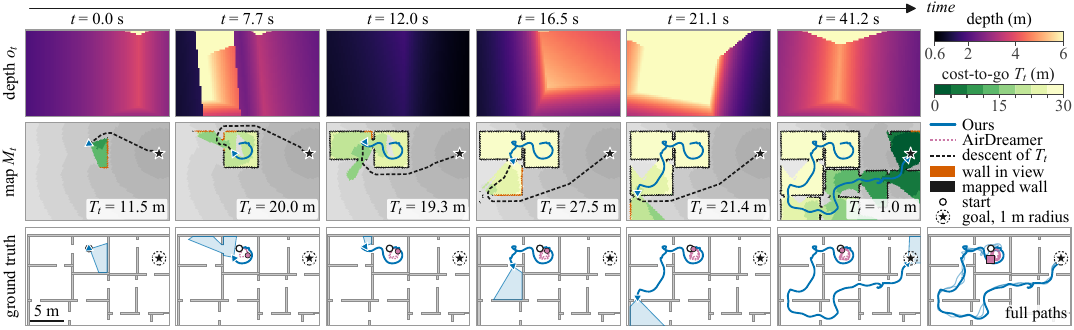}
\caption{A test flight of \ours{} through a 16-room building of the in-distribution suite ($R_g=3.49$). The rows show
the depth image $o_t$, the scan-memory map $M_t$ built so far, and the ground truth with the camera's field of view.
The map $M_t$ is colored by the cost-to-go $T_t$ and gray where not yet observed, and the dashed line follows the
steepest descent of $T_t$. At 7.7, 12.0, and 16.5\,s, a newly mapped wall has just closed the route. The right column
shows the full paths of the three runs of \ours{}, two of them pale, and of one AirDreamer run after 1.0M steps,
whose end is marked by a square.}
\label{fig:timeline}
\end{figure*}

The gap grows with the detour ratio $R_g$ (Fig.~\ref{fig:detour}). For $R_g\ge2$, \ours{} succeeds in
\msd{94.2}{0.6}\% of the episodes and never times out, whereas AirDreamer succeeds in \msd{22.9}{4.4}\% at the same
budget and in \msd{26.2}{10.1}\% after 2.0M steps. In the C-shaped cavity of Fig.~\ref{fig:suites},
\ours{} solves 49 or 50 of the 50 scenes, and AirDreamer solves 17 to 27 at the same budget.

Fig.~\ref{fig:curves} shows that \ours{} also learns faster per environment step. After 0.8M steps,
\msd{90.1}{2.9}\% of its training episodes succeed, against \msd{51.8}{2.7}\% for AirDreamer at the same budget and
\msd{61.8}{9.2}\% after 2.0M steps.

\begin{figure}[t]
\centering
\includegraphics[width=\columnwidth]{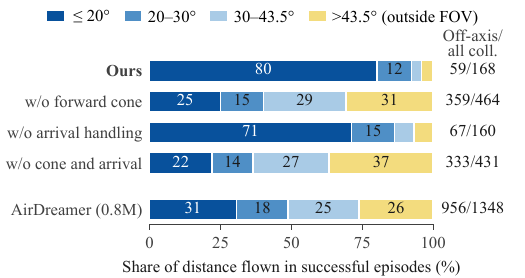}
\caption{Direction of horizontal motion relative to the camera axis at 0.8M steps, as shares of the distance flown
in successful test episodes. The right column gives the off-axis collisions among all collisions.}
\label{fig:fwd}
\end{figure}

AirDreamer's failures match the first three difficulties of Sec.~\ref{sec:intro}. With its memory held only in the
recurrent state, 482 of its 527 timeouts at our budget are loops, in which the quadrotor circles or shuttles over
ground it has already covered. In 53.1\% of its timeouts, it had closed more than 20\% of the straight-line distance
to the goal, which the Euclidean progress term rewards, but less than 20\% of the geodesic distance. Since its reward
ignores the camera direction, it flies \msd{26.2}{10.3}\% of its distance in successful episodes outside the FOV, and
956 of its 1348 collisions are off-axis (Fig.~\ref{fig:fwd}).

The map also lets \ours{} give up a route that turns out to be closed. In the flight of Fig.~\ref{fig:timeline}
through a 16-room building, a newly mapped wall closes the route three times, so the cost-to-go jumps within one step
and the path turns toward another door. These routes stay closed after their walls leave the view. All runs of
\ours{} reach the goal, whereas AirDreamer stays in the start room at all three budgets. Of the 100
in-distribution multi-room scenes, \ours{} solves 81 to 91 and AirDreamer 21 to 38 at the same budget.

The released NavRL policy solves every open-space scene but almost no scene with a large detour
(Fig.~\ref{fig:detour}). Of its 1032 failures, 850 are timeouts, 793 of them loops. With its yaw held on the
start-goal line, it flies detours sideways or backward, and 172 of its 177 collisions are off-axis. The retrained
NavRL and TD-MPC2 succeed even less often, and neither solves any episode with $R_g\ge2$, but they fail in different
ways: the retrained NavRL hovers in place in 1006 of its 1055 timeouts after 2.0M steps, whereas TD-MPC2 collides in
about 52\% of the episodes at every budget, mostly off-axis (659 of 819 collisions after 2.0M steps).

\subsection{Component Analysis}
\label{sec:ablation}

To measure what each component contributes, we remove components from \ours{} and train the variants with the same
setup (Table~\ref{tab:ablation}). The last two variants, which lack both the forward-cone penalty and arrival
handling, also use a 2D potential without $\Delta z_t$.

\subsubsection{Forward-Cone Penalty}
Removing the penalty lowers success from \msd{96.3}{0.3}\% to \msd{89.9}{3.1}\% and raises collisions from
\msd{3.6}{0.3}\% to \msd{9.9}{3.1}\%. Without it, the forward-motion share drops from 92.4\% to 40.1\%
(Fig.~\ref{fig:fwd}), and the variant alone collides off-axis in 339 episodes, against 39 for \ours{} alone. Success
falls from 94.4\% to 83.9\% on the topology-shift suite, from 89.5\% to 79.5\% on the parameter-shift suite, and from
97.0\% to 90.1\% on the structured suite, where all 89 failures are collisions, 67 of them in the narrow passage.

\subsubsection{Near-Goal Arrival Handling}
Arrival handling has no consistent effect on success or collision (Table~\ref{tab:ablation}) but changes how
\ours{} arrives. Its final approach takes \msd{0.38}{0.01}\,s instead of \msd{1.39}{0.30}\,s and is shorter in 3755
of the episodes both solve and longer in 288.

\subsubsection{Both Components}
Removing both components lowers success to \msd{90.3}{3.3}\% and raises collisions to \msd{9.2}{3.1}\%. This
variant succeeds about as often as the one without the cone, but 13 of its 24 timeouts are the near-goal stalls that
arrival handling targets (Sec.~\ref{sec:imag}), whereas the variant without the cone, which keeps arrival handling
and the 3D potential, has none.

\subsubsection{Geodesic Reward Only}
By itself, the geodesic reward does not help: replacing the Euclidean term of AirDreamer, it lowers success from
\msd{59.6}{3.0}\% to \msd{51.6}{2.5}\% at the same budget. Once the map and map-grounded imagination are added, as in
the variant without the cone and arrival handling, success rises to \msd{90.3}{3.3}\%. We hypothesize that this reward
needs the map as an input, since otherwise a newly mapped wall changes it in a way the networks cannot see.

% \section{Limitations and Future Work}
% \label{sec:limitations}

\section{Conclusion}
\label{sec:conclusion}

This paper presented \ours{}, a map-grounded world-model agent for
quadrotor navigation with a forward-facing depth camera. Using the
quadrotor pose, \ours{} constructs an online map that provides spatial
memory, defines a geodesic-progress reward, and grounds imagined
rollouts. \ours{} also penalizes horizontal motion outside a forward cone
around the camera axis. On unseen simulated layouts, it learns faster per
environment step than AirDreamer, reaches the goal far more often, and has
fewer collisions, even when AirDreamer trains longer. Our results suggest
that world-model agents do better when they build in the geometric and
sensing structure a robot already has (its pose, what it has observed so
far, and where its sensor points) instead of learning all of it in a
latent state.

Future work will first address more diverse and more realistic
environments. Our 2-D map suits obstacles that span the flight band,
whereas real buildings and subterranean sites~\cite{tranzatto2022cerberus}
also contain overhanging obstacles that call for a 3-D map. On a real
quadrotor, the map must be built from noisy depth images and a pose from
visual-inertial odometry. We will then address dynamic goals and obstacles.

\IEEEtriggeratref{14}
\bibliographystyle{IEEEtran}
\bibliography{refs}

\end{document}